\documentclass{article}

\usepackage{float}
\usepackage[preprint]{corl_2026} 

\usepackage{amsmath}
\usepackage{amssymb}
\usepackage{bm}
\usepackage{graphicx}
\usepackage{array}
\usepackage{booktabs}
\usepackage{xcolor}
\usepackage{multirow}
\newcommand{\df}{\mathrm{d}}

\newtheorem{problem}{Problem}

\newcommand{\best}[1]{\textbf{#1}}
\newcommand{\second}[1]{\underline{#1}}
\title{\textbf{PLATO}: \emph{Preintegration Learning from Accurate Trajectory Observations} for Neural Inertial Odometry}
\author{
Haoying Li, Qihang Liu, Yifan Peng, Keyan Miao, and Junfeng Wu 
}

\begin{document}
\maketitle

\begin{abstract}
Neural inertial odometry has demonstrated strong potential for motion estimation in challenging environments, yet inertial-only preintegration remains sensitive to IMU bias and uncertainty.
To this end, this paper introduces \textbf{PLATO}:~\emph{Preintegration Learning from Accurate Trajectory Observations}, a likelihood-based framework that leverages accurate trajectory observations to jointly learn IMU bias dynamics modeled by a neural ordinary differential equation~(NODE) and gyroscope and accelerometer noise covariances.
Optimization exploits the sparse structure of the negative log-likelihood, with IMU noise-parameter gradients computed by forward differentiation.
A tailored double-adjoint scheme couples a discrete invariant-error adjoint with a continuous-time adjoint for the bias NODE, enabling memory-efficient likelihood optimization over the nested bias-dynamics and IMU-preintegration rollouts.
Validation on EuRoC shows improved performance, and underwater robot experiments demonstrate applicability under intermittent lighting failures and visual degradation.

\end{abstract}


\section{Introduction}
Inertial measurement units~(IMUs) are low-cost, low-power proprioceptive sensors that provide high-rate measurements of a robot's angular velocity and linear acceleration, which are insensitive to external environmental conditions.
However, correctly using IMU measurements requires accounting for accelerometer and gyroscope biases, whose misspecifications accumulate rapidly through high-rate integration and soon manifest as pose-estimation drift.
Although most modern inertial odometry systems estimate IMU biases online~\cite{geneva2020openvins,10757429}, reliable bias correction depends on exteroceptive sensor constraints. Thus, when such sensors fail, the estimator will rapidly degrade.

Learning-based methods have shown strong potential to better leverage IMU measurements and improve system robustness.
Existing neural inertial odometry~(NIO) methods can be broadly grouped into the following two lines.
The first type is called neural displacement priors~(NDPs), which learn to map raw IMU measurements to displacement~\citep{jayanth2025neural}. 
RoNIN learns the 2D motion of pedestrians from raw IMU sequences~\cite{herath2020ronin}.
TLIO regresses 3D displacement and covariance using Residual Network~(ResNet) from a buffer of IMU measurements, and tightly fuses the learned relative-motion measurements in an extended Kalman filter~(EKF)~\cite{liu2020tlio}.
IDOL improves pedestrian localization by learning device orientation using a recurrent neural network~\cite{sun2021idol}.
Another line of work employs neural networks to debias IMU measurements.
\citet{brossard2020denoising} use past gyroscope and accelerometer measurements to learn gyroscope corrections and use them for open-loop orientation estimation. 
AirIMU learns IMU correction and uncertainty propagation through a differentiable inertial integrator~\cite{qiu2023airimu}. 
\citet{zhou2025learning} model IMU bias using a diffusion model, emphasizing its stochastic nature. 
\citet{altawaitan2025learned} predict IMU biases from inertial histories by minimizing the discrepancy between bias-corrected preintegrated states and ground-truth states, and then use the estimated biases within an invariant visual-inertial odometry~(VIO).
Rather than directly mapping local inertial-measurement windows to instantaneous bias estimates, some works advocate learning the temporal evolution of IMU biases. \citet{buchanan2022deep} learn the IMU bias evolution model from the ground-truth bias labels, and illustrate that the dynamics-based formulation encourages the network to capture sensor-intrinsic drift behavior instead of motion-specific correlations, thereby improving generalization to unseen motion patterns.
Recent work~\cite{liu2025debiasing} models IMU bias evolution using a neural ordinary differential equation~(NODE) from ground-truth poses.

\begin{figure}[t]
    \centering
    \includegraphics[width=1.0\linewidth]{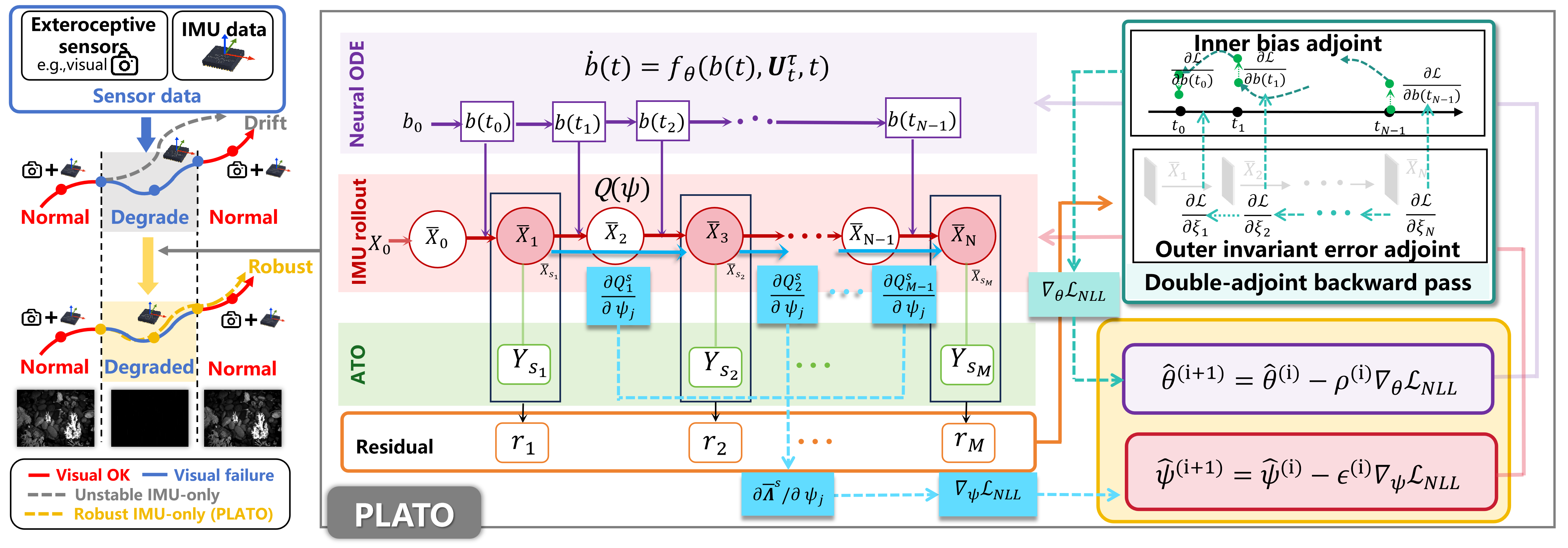}
    \caption{PLATO framework for ATO-supervised neural inertial preintegration.}
    \label{fig:overview}
\end{figure}

This paper learns IMU bias dynamics and measurement uncertainty to refine the IMU measurement model,  which is more interpretable than directly learning the full IMU-to-pose mapping from data.
The most closely related work is~\cite{liu2025debiasing}, which trains the bias NODE by minimizing the mean-squared error~(MSE) between debiased IMU rollout trajectories and ground-truth poses. 
In contrast, this paper casts training as likelihood maximization, enabling
joint optimization of bias dynamics and noise statistics while exploiting
the problem structure for efficient gradient computation.
To this end, this paper proposes \textbf{PLATO}:
\textbf{P}reintegration \textbf{L}earning from
\textbf{A}ccurate \textbf{T}rajectory \textbf{O}bservations
for neural inertial odometry, whose overview is shown in
Fig.~\ref{fig:overview}.
PLATO treats trajectory information from different sources, such as
motion-capture ground truth and visual odometry estimates, uniformly as
accurate trajectory observations~(ATOs).
By explicitly accounting for their observation models and uncertainties,
PLATO incorporates these observations into a unified marginal
negative log-likelihood objective for jointly learning IMU bias dynamics
and noise statistics.
The learned IMU model then enables more reliable inertial-only
preintegration when external trajectory observations become degraded
or unavailable.
The contributions of this paper are summarized as follows:
\begin{itemize}
\item \textbf{Sparse marginal-likelihood training from ATOs.}
The inertial trajectory is treated as a sequence of latent variables driven by NODE-based IMU bias dynamics.
The resulting marginal-likelihood enables joint learning of bias dynamics and IMU noise statistics from ATO supervision, supporting ground-truth trajectories and noisy pose observations.

\item \textbf{Structured derivative computation for efficient training.}
Efficient derivatives are derived by exploiting the sparse negative log-likelihood and IMU rollout structure.
The noise-parameter update uses forward differentiation through the covariance terms in the information-form likelihood.
For bias-dynamics training, a dedicated double-adjoint scheme is developed.
By coupling an outer invariant-error adjoint with an inner bias adjoint, the scheme avoids direct reverse-mode automatic differentiation over the full nested rollout spanning both NODE solver and IMU propagation steps.

\item \textbf{Validation on public and self-collected underwater datasets.}
Experiments on the EuRoC and the self-collected underwater robot dataset \textsc{AquaLux} demonstrate improved inertial propagation and better downstream OpenVINS performance. 
\textsc{AquaLux} will be released as a benchmark for inertial-only odometry under intermittent visual failures.
\end{itemize}

\section{Preliminaries}
\subsection{Matrix Lie Group}
The Lie group and Lie algebra preliminaries~\cite{sola2018micro} used throughout this paper are briefly reviewed.
Let $\mathbf{G}$ be a matrix Lie group with Lie algebra $\mathfrak{g}$. 
We abuse the notation $(\cdot)^\wedge$ to denote the mapping from the vector space to the corresponding Lie algebra with its inverse map $(\cdot)^\vee$.
The exponential map is defined as $\operatorname{exp}(\xi)\triangleq \operatorname{exp}_m(\xi^\wedge): \mathfrak{g} \rightarrow G$, where $\exp_m(\cdot)$ denotes the matrix exponential. 
The inverse mapping of the exponential is the Lie logarithm, denoted by $\log(\cdot)$.
The special orthogonal group $SO(3)$ represents the set of all possible rotations of a rigid body in three-dimensional space. 
The special Euclidean group $SE(3)$~(rigid-body transformations), and $SE_2(3)$ are defined as
{\small
\[
\mathrm{SE}(3)\triangleq\left\{
\left[\begin{array}{cc}
R & p\\
0 &1
\end{array}\right]\;\middle|\;
R\in \mathrm{SO}(3),\ p\in\mathbb R^{3}\right\},
\quad
\mathrm{SE}_2(3) \triangleq\left\{
\left[
\begin{array}{cc}
R & v \;\; p \\ 
\bm 0 & I
\end{array}
\right]
\middle|
R \in \mathrm{SO}(3),\ p,v \in \mathbb{R}^3
\right\}.
\]
}
where $\bm 0$ and $I$ denote the zero matrix and the identity matrix.
For $x^{\wedge}, y^{\wedge} \in \mathfrak{g}$ and $\|x\|$ small, their compounded exponentials can be approximated to
$
 \exp (x) \exp (y) \approx \exp ( \operatorname{dexp}_{y}^{-1} x+y),
$
where $\operatorname{dexp}_x$ is the left Jacobian of $x$. 
Let $\mathcal{M}$ be the smooth manifold of a Lie group.  
For any $X_1, X_2 \in \mathcal{M}$ and tangent vector $\xi \in \mathbb{R}^{\dim \mathcal{M}}$,  define the \emph{plus} and \emph{minus} operators, $X_1 \boxplus \xi$ and $X_1 \boxminus X_2$, and in the left case,
$
 X_1 \boxplus \xi =  \exp(\xi){X}_1, X_1  \boxminus X_2 = \log(X_1X_2^{-1}).
$
The uncertainty of $X \in \mathcal{M}$ is modeled in the tangent space by a Gaussian distribution.  
Let $\xi \sim \mathcal{N}(0,\Sigma)$, the induced distribution is
$
 \mathcal{N}_L(\bar{X}, \Sigma) \triangleq 
 \alpha \exp\left(-\tfrac{1}{2} (X \boxminus \bar{X})^\top \Sigma^{-1} (X \boxminus \bar{X})\right),
$
and $\alpha$ is the normalization coefficient.
Hereafter, $X\in \mathrm{SE}_2(3)$ denotes a state, whereas $T\in \mathrm{SE}(3)$ denotes a pose.

\section{Problem Formulation}
\label{sec:problem_formulation}
IMU measurements are commonly incorporated through the following rigid-body kinematics:
\begin{equation}\label{eq:imukin}
    \dot{R}(t) = R(t)~\omega(t)^\wedge~,\quad \dot{p}(t)  = v(t), \quad \dot{v}(t) = a(t),
\end{equation}
in which the measurement model of IMU is given by
\begin{equation}\label{eq:IMUmeasurement}
   \tilde \omega(t) = \omega(t) + b_g(t) + n_g(t), \quad \tilde{a}(t) = R(t)^\top (a(t)-g) + b_a(t) + n_a(t),
\end{equation}
where \(\omega(t)\) and \(a(t)\) denote the angular velocity and linear acceleration of the rigid body, 
$g$ denotes the gravity,
$b_g(t)$ and $b_a(t)$ denote the gyroscope and accelerometer bias, 
$n_g(t)$ and $n_a(t)$ are the corresponding white Gaussian noises, and $\tilde \omega(t)$, $\tilde a(t$) are the IMU measurements.
Denote the IMU measurement sequence ${\bm U}=\{u_0,\ldots,u_N\}$ with $u_k = (\tilde{a}_k,\tilde{\omega}_k)$ obtained at timestamp $t_k$.
Define $\bm U^{\tau}_t$ as the collection of discrete IMU measurements from the time interval $( t-\tau, t],\tau>0$.
The initial state \(X_0\in SE_2(3)\) and the initial bias \(b_0=\operatorname{vec}(b_0^a,b_0^g)\) are provided by the initialization procedure. 
The IMU biases are modeled as deterministic variables governed by a NODE as follows: 
\begin{equation}\label{eq:bias_dynamics}
    \dot b(t) = f_\theta\bigl(b(t), \bm U_t^{\tau} ,t\bigr), \qquad b(t_0)=b_0,
\end{equation}
where \(\theta\) denotes the neural dynamics parameters.
Let \(\psi\) specify the covariance of the accelerometer and gyroscope noises, $Q(\psi) = \operatorname{cov}(n)$ and $n=\operatorname{vec}(n_a,n_g)$.
All the learnable parameters are collectively denoted as $\Theta \triangleq (\theta,\psi)$.
Let $\Omega$ denote the admissible parameter set, which encodes constraints such as boundedness and positivity of the noise covariance parameters.
During training, a set of ATOs is available, denoted by $\bm Y^s \triangleq \{Y_{s_1},Y_{s_2},\cdots, Y_{s_M}\}$, sampled at time $t_{s_j}, j=1,\cdots,M$.
The overall parameter-learning problem is thus formulated as follows:
\begin{problem}\label{prob:1}
Given the IMU sequence $\bm U$, the ATO sequence $\bm Y^s$, and the initial conditions $X_0$ and $b_0$, estimate the admissible parameters $\Theta\in\Omega$ by maximizing the marginal likelihood
\begin{equation} \label{eq:parameter_learning_objective}
    \hat\Theta
    =
    \arg\max_{\Theta\in\Omega}
    p\bigl(
       \bm Y^s
        \mid
       \bm U,X_0,b_0,\Theta
    \bigr).
\end{equation}
\end{problem}

To solve Problem~\ref{prob:1}, a latent trajectory is introduced as the collection of states that are supervised by ATOs, represented as $\bm X^s\triangleq\{X_{s_1},\cdots, X_{s_M}\}$.
Marginalizing over the latent trajectory gives:
\begin{equation}
    p\bigl(\bm Y^s
        \mid \bm U,X_0,b_0,\Theta
    \bigr)
    =
    \int
    p(\bm Y^s \mid \bm X^s)
    p(\bm X^s\mid \bm U,X_0,b_0,\Theta)
    \df \bm X^s .
    \label{eq:marginal_likelihood_integral}
\end{equation}

In the following, Section~\ref{sec:likelihood_evaluation} presents a sparsity-exploiting likelihood evaluation strategy induced by the IMU rollout, 
and Section~\ref {sec:Optimization} provides an efficient derivative computation scheme for training.

\section{Likelihood Evaluation from Pure IMU Rollout}
\label{sec:likelihood_evaluation}

\subsection{Pure IMU Rollout as a Latent Trajectory}
\label{subsec:imu_rollout_latent}

The pure IMU preintegration is used to compute $p(\bm X^s\mid \bm U,X_0,b_0,\Theta)$ required in~\eqref{eq:marginal_likelihood_integral}.
The bias is obtained by integrating \eqref{eq:bias_dynamics}, and the bias-corrected inertial inputs are then used to propagate a nominal motion trajectory according to~\eqref{eq:imukin}, represented by
$     \bar{\bm X}
    =
    \{\bar X_0,\bar X_1,\ldots,\bar X_N\},
    \bar X_k\in SE_2(3)$.
Define the right-invariant error \(\eta_k=X_k\bar X_k^{-1}\) and its logarithmic coordinate \(\xi_k=\log(\eta_k)\). 
As shown by~\cite{barrau2016invariant}, for the group-affine inertial kinematics in~\eqref{eq:imukin}, the deterministic part of the right-invariant error dynamics is independent of the nominal state.
Accordingly, the propagated uncertainty is evaluated in the corresponding Lie algebra and the first-order discretization is as follows
\begin{equation}\label{eq:IMUXP}
\bar P_{k+1}=F_k\bar P_kF_k^\top+G_kQ(\psi)G_k^\top,
\qquad
n_k\sim\mathcal N(0,Q(\psi)),
\end{equation}
where  \(\Delta t_k=t_{k+1}\!-\!t_k\),
\(F_k=\Phi(\Delta t_k)=\exp_m(A\Delta t_k)\), with $A=
\left[
\begin{smallmatrix}
0 & 0 & 0\\
g^\wedge & 0 & 0\\
0 & I & 0
\end{smallmatrix}
\right]$, and 
$G_k$
is the first-order noise-input matrix evaluated along the nominal rollout~\cite{hartley2020contact}. 
The pure IMU preintegration induces an approximate distribution $
    p\bigl(
        \bm X^s
        \mid
      \bm  U,X_0,b_0,\Theta
    \bigr)
    \approx
    \mathcal N_L
    \bigl(
        \bar{\bm X}^s,\bar{\bm P}^s
    \bigr)$,
where the joint covariance \(\bar{\bm P}^s\) is induced by the corresponding linear Gaussian Markov chain and includes the cross-covariances among retained states.

\subsection{Marginal Likelihood Evaluation}
\label{subsec:general_marginal_likelihood}

This subsection presents the sparse information-form evaluation of the marginal likelihood. 
Although \(\bar{\bm P}^s\) is generally dense, the information matrix \(\bar{\bm\Lambda}^s=(\bar{\bm P}^s)^{-1}\) is block-tridiagonal due to the linear Gaussian Markov structure of inertial preintegration~\cite{barfoot2014batch}. 
To make this sparse precision explicit, for \(r\ge q\), define
$
    \Phi_{r\leftarrow q}
    \triangleq
    F_{r-1}F_{r-2}\cdots F_q,~
    \Phi_{q\leftarrow q}=I .
$
For two consecutive supervised time index $s_i$ and $s_{i+1}$, define
$
    \Phi_i^s
    \triangleq
    \Phi_{s_{i+1}\leftarrow s_i}
$
and the accumulated process covariance
\begin{equation}\label{eq:Qis1}
    Q_i^s
    \triangleq
    {\textstyle\sum_{k=s_i}^{s_{i+1}-1}}
    \Phi_{s_{i+1}\leftarrow k+1}
    G_k Q(\psi)G_k^\top
    \Phi_{s_{i+1}\leftarrow k+1}^\top .
\end{equation} 
The nonzero blocks of \(\bar{\bm\Lambda}^s\) are
\(\bar{\Lambda}_{11}^s=(P_1^s)^{-1}+(\Phi_1^s)^\top(Q_1^s)^{-1}\Phi_1^s\),
\(\bar{\Lambda}_{MM}^s=(Q_{M-1}^s)^{-1}\),
\(\bar{\Lambda}_{ii}^s=(Q_{i-1}^s)^{-1}+(\Phi_i^s)^\top(Q_i^s)^{-1}\Phi_i^s\) for \(i=2,\ldots,M-1\), and
\(\bar{\Lambda}_{i,i+1}^s=-(\Phi_i^s)^\top(Q_i^s)^{-1}\),
\(\bar{\Lambda}_{i+1,i}^s=-(Q_i^s)^{-1}\Phi_i^s\) for \(i=1,\ldots,M-1\), where \(P_1^s\succ0\) denote the prior covariance.

For an ATO \(Y_{s_i}\) associated with the state \(X_{s_i}\), the residual is defined as
$r_i(\Theta) =Y_{s_i}\boxminus h_i(\bar X_{s_i}).$
Let \(H_i\) be the Jacobian of the residual with respect to the invariant error, and $v_i \sim \mathcal{N}(\bm 0, W_i)$ denotes the Gaussian measurement noise.
Stacking all residuals gives
$\bm r=\operatorname{vec}(r_1,\ldots,r_M)$, $H=\operatorname{blkdiag}(H_1,\ldots,H_M)$ and 
$\bm W=\operatorname{blkdiag}(W_1,\ldots,W_M)$.
Under the Gaussian approximation, the stacked residual is distributed as
\(\bm r\sim\mathcal N(\bm 0,\bm S)\), where
\(\bm S=\bm H\bar{\bm P}^s\bm H^\top+\bm W\).
Up to additive constants independent of \(\Theta\), the corresponding negative log-likelihood~(NLL) is 
\begin{equation}\label{eq:generalnll}
    \mathcal{L}_{\mathrm{NLL}}(\Theta)
    =
    \frac12
    \bm r^\top \bm S^{-1}\bm r
    +
    \frac12
    \log\det \bm S.
\end{equation}

This general NLL~\eqref{eq:generalnll} is specialized to two ATO settings: noise-free and noisy pose supervision. 
The noise-free setting uses ground-truth poses, such as those from motion-capture-covered segments, to train the network for robust inertial propagation when such supervision is unavailable.
Since the supervision noise covariance is set to \(\bm W=\bm 0\), one has \(\bm S=\bm H\bar{\bm P}^s\bm H^\top\), thus \(\bm H\) is required to be locally invertible. Accordingly, ground-truth pose observations are lifted to full states on \(\mathrm{SE}_2(3)\), with velocities estimated by finite differences of positions.
With \(\tilde{\bm r}=\bm H^{-1}\bm r\), the NLL reduces to
\begin{equation}\label{eq:gtnll}
        \mathcal L_{\mathrm{NLL}}^{\mathrm{gt}}(\Theta)
    =
    \frac12\tilde{\bm r}^\top\bar{\bm\Lambda}^s\tilde{\bm r}
    -
    \frac12\log\det\bar{\bm\Lambda}^s
    +
    \log|\det\bm H| .
\end{equation}

For noisy pose supervision, the ATO residuals are modeled with a positive-definite supervision covariance \(\bm W\succ\bm 0\), and the marginal NLL is evaluated in sparse information form~\citep{barfoot2014batch}. Let \(\bm K\triangleq\bar{\bm\Lambda}^s+\bm H^\top\bm W^{-1}\bm H\), \(\bm\gamma\triangleq\bm H^\top\bm W^{-1}\bm r\), and \(\bm K\bm z=\bm\gamma\). The resulting NLL is
\begin{equation}\label{eq:vonll}
    \mathcal{L}_{\mathrm{NLL}}^{\mathrm{pose}}(\Theta)
    =
    \frac{1}{2}\bm r^\top\bm W^{-1}\bm r
    -
    \frac{1}{2}\bm\gamma^\top\bm z
    +
    \frac{1}{2}\log\det\bm W
    +
    \frac{1}{2}\log\det\bm K
    -
    \frac{1}{2}\log\det\bar{\bm\Lambda}^s .
\end{equation}

Both NLL evaluations can be computed efficiently in sparse information form, without explicitly forming \(\bar{\bm P}^s\) or \(\bm S\); details are in Appendix~\ref{sec:sparse_nll}.
The next section exploits the rollout and likelihood structures to derive efficient gradients for the bias-dynamics and noise-covariance parameters.

\section{Optimization Procedure}
\label{sec:Optimization}
\subsection{Overall Training Strategy}

Before optimizing $\mathcal L_{\mathrm{NLL}}$, the bias dynamics parameters $\theta$ are first trained with the MSE objective $\mathcal L_{\mathrm{MSE}}(\theta)=\frac12\sum_{i=1}^{M}\|r_i(\theta)\|_2^2$.
This serves as the warm-up for the subsequent NLL optimization, and stabilizes the bias trajectory mean before covariance parameters are optimized, preventing the use of covariance inflation to absorb large residuals~\citep{liu2020tlio,buchanan2022learning}.

\begin{figure}[H]
    \centering
    \includegraphics[width=0.8\linewidth]{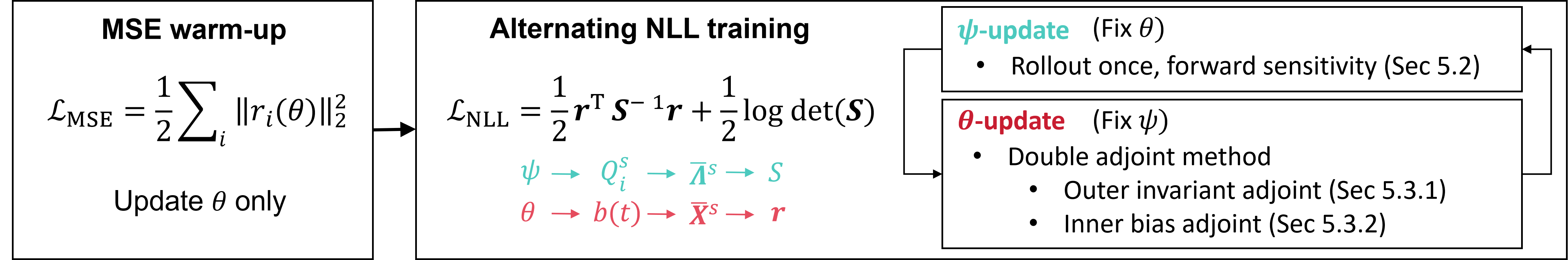}
    \caption{Training strategy and structured gradient computation.}
    \label{fig:train}
\end{figure}

After the MSE warm-up, training switches to the marginal NLL and alternates between the two parameter blocks, as summarized in Fig.~\ref{fig:train}. The following subsections derive the \(\psi\)-update based on forward covariance sensitivities and the \(\theta\)-update based on the double-adjoint method.
Efficient gradient computations for $\psi$ and $\theta$ are derived next and are previewed in Figure~\ref{fig:train}.

\subsection{Forward Sensitivity for Covariance Training}
During the optimization of \(\psi\), the bias-dynamics parameters \(\theta\) are fixed.
Consequently, the nominal states, the residuals, and the linearized propagation
matrices \(F_k\) and \(G_k\) are fixed.
The dependence of \(\mathcal L_{\mathrm{NLL}}\) on \(\psi\) is therefore only
through the IMU noise covariance \(Q(\psi)\), the induced interval covariances
\(Q_i^s(\psi)\), and the corresponding precision blocks in
\(\bar{\bm\Lambda}^s\).
For each supervisory interval \([s_i,s_{i+1}]\), the interval covariance~\eqref{eq:Qis1} can be
computed by a local covariance recursion
\begin{equation}\label{eq:Qs_recursion1}
        \Sigma_{i,k+1}
    =
    F_k\Sigma_{i,k}F_k^\top
    +
    G_kQ(\psi)G_k^\top,
    \qquad k=s_i,\ldots,~s_{i+1}\!-\!1,
\end{equation}
where \(\Sigma_{i,s_i}=\bm 0\) and the recursion gives that \(Q_i^s=\Sigma_{i,s_{i+1}}\).
Then differentiating~\eqref{eq:Qs_recursion1} term by term:
\begin{equation}\label{eq:dQs_recursion}
    \frac{\partial \Sigma_{i,k+1}}{\partial \psi_j}
    =
    F_k
    \frac{\partial \Sigma_{i,k}}{\partial \psi_j}
    F_k^\top
    +
    G_k
    \frac{\partial Q(\psi)}{\partial \psi_j}
    G_k^\top,
    \qquad k=s_i,\ldots,~s_{i+1}\!-\!1,
\end{equation}
where $    \frac{\partial \Sigma_{i,s_i}}{\partial \psi_j}=\bm 0,
    \;
    \frac{\partial Q_i^s}{\partial \psi_j}
    =
    \frac{\partial \Sigma_{i,s_{i+1}}}{\partial \psi_j}$.
The interval-covariance sensitivities are assembled to obtain
\(\partial\bar{\bm\Lambda}^s/\partial\psi_j\) in subsection~\ref{subsec:general_marginal_likelihood}, preserving the same sparse
block-tridiagonal structure. 
Therefore, by the chain rule, the derivative w.r.t $\psi$ is computed as
$\frac{\partial \mathcal L_{\mathrm{NLL}}}{\partial \psi_j}
=
\left\langle
\frac{\partial \mathcal L_{\mathrm{NLL}}}{\partial \bar{\bm\Lambda}^s},
\frac{\partial \bar{\bm\Lambda}^s}{\partial \psi_j}
\right\rangle$.
Due to the low dimension of $\psi$ and the fixed \(F_k\), \(G_k\), these forward sensitivities require a small number of additional covariance recursions and do not store the high-rate covariance rollout in a reverse-mode computation graph, leading to a more time- and memory-efficient update than direct reverse-mode AD.

\subsection{Double Adjoint Method for Bias-Dynamics Training}
The \(\theta\)-update requires differentiating the NLL through the bias NODE and the induced inertial preintegration rollout. Direct reverse-mode AD stores the discretized computation graph, whose memory cost scales as \(\mathcal O(JL)\), where \(J\) denotes the activation memory per rollout step and \(L\) is the total number of discretized steps. In contrast, adjoint methods propagate loss sensitivities backward through the latent dynamics and reduce the memory cost to \(\mathcal O(L)\)~\cite{kidger2020neural}.
In our problem, the continuous NODE integration that generates the bias trajectory is embedded within high-rate inertial preintegration. Consequently, direct reverse-mode AD constructs a nested long-horizon computation graph involving both the NODE solver trajectory and the IMU propagation rollout. 
To this end, a double-adjoint method is developed: the outer invariant-error adjoint provides bias-level sensitivities from inertial preintegration, which are incorporated by the inner bias adjoint for NODE training.

\subsubsection{Outer Invariant Error Adjoint for Bias Sensitivity}
\label{subsec:invariant_adj}

The outer adjoint characterizes the sensitivity of the outer loss with respect
to the queried IMU bias values. This formulation is natural because, under the
group-affine inertial dynamics, the invariant error admits a state-independent
linearized propagation. Let \(b_k^\star\) denote the true IMU bias. Define the
bias error $\tilde b_k = b_k-b_k^\star$.
Discretizing the invariant-error dynamics in~\cite{li2022closed} gives
\begin{equation}\label{eq:dis}
    \xi_{k+1}
    =
    F_k\xi_k
    +
    G_k(n_k-\tilde b_k),
\end{equation}
where \(F_k\) and \(G_k\) are evaluated along the nominal rollout.
For clarity, a general loss $\mathcal L$ is temporarily introduced.
Applying the chain rule to~\eqref{eq:dis}, starting from
$\lambda_N = \frac{\partial \mathcal L}{\partial \xi_N}$, the adjoint method satisfies the backward recursion
\begin{align}
\lambda_k
&=
\frac{\partial \mathcal L}{\partial \xi_k} + F_k^\top \lambda_{k+1},\quad \frac{\partial \mathcal L}{\partial b_k}
=
\left(
\frac{\partial \xi_{k+1}}{\partial b_k}
\right)^\top
\lambda_{k+1}
=
- G_k^\top \lambda_{k+1}, 
\quad k = N-1,\ldots,1.
\label{eq:adjoint}
\end{align}

Thus, the outer adjoint converts loss sensitivity on inertial invariant errors
into discrete sensitivities for queried bias values, which serve as inputs to
the inner NODE adjoint for computing $\partial \mathcal L/\partial \theta$.

\subsubsection{Inner Adjoint Method for Bias NODE Training}
\label{subsec:adjoint bias}

Let \(b_k \triangleq b(t_k)\) be the bias queried at the IMU timestamp \(t_k\).
The sequence \(\{b_k\}_{k=0}^{N}\) forms an intermediate latent trajectory between the NODE parameter \(\theta\) and the loss \(\mathcal L\).
The outer invariant error adjoint yields discrete gradients with respect to the queried bias values at the nominal rollout
$
    g_k^b=\frac{\partial \mathcal L}{\partial  b_k}
$.
These gradients are incorporated through the inner adjoint
$
    \lambda_b(t)
    \triangleq
    \frac{\partial \mathcal L}{\partial b(t)} .
$
The inner adjoint is propagated backward and updated at IMU query times as
$
\lambda_b(t_k^-)
=
\lambda_b(t_k^+)+g_k^b .
$
Between query times, it satisfies~\cite{chen2018neural}:
\begin{equation}
    \dot \lambda_b(t)
    =
    -
    \left(
    \frac{\partial f_\theta}{\partial b}
    \bigl(b(t), \bm U_t^{\tau}, t\bigr)
    \right)^\top
    \lambda_b(t),
    \qquad t\in(t_{k-1},t_k).
    \label{eq:inner_adjoint_dynamics}
\end{equation}

The gradient with respect to the NODE parameter is then accumulated along the continuous bias dynamics as
$
    \frac{d\mathcal{L}}{d\theta}
    =
    \int_{t_0}^{t_N}
    \left(
    \frac{\partial f_\theta}{\partial \theta}
    \bigl(b(t), \bm U_t^{\tau}, t\bigr)
    \right)^\top
    \lambda_b(t)
    \, dt .
$
Thus, the inner adjoint converts outer-adjoint bias sensitivities into the NODE parameter gradient, yielding a double-adjoint mechanism for learning IMU bias dynamics.
The overall framework is summarized in Fig.~\ref{fig:overview}.

\section{Experiments}
\label{sec:experiments}

This section evaluates the practical effectiveness of the proposed method on public and self-collected datasets.
The experiments examine two aspects: raw IMU preintegration after debiasing and downstream OpenVINS~\cite{geneva2020openvins} performance. 
The evaluation metrics are provided by~\cite{grupp2017evo}, including absolute orientation error~(AOE) and absolute position error~(APE), defined as
\(\mathrm{AOE}=(\sum_{k=1}^N\|\log(\hat R_k^\top R_k)\|^2/N)^{1/2}\)
and
\(\mathrm{APE}=(\sum_{k=1}^N\|p_k-\hat p_k\|^2/N)^{1/2}\).
The baselines include: 1) using raw IMU data;
2) Hierarchical Learning of Continuous Bias Dynamics (HL-CBD)~\cite{liu2025debiasing};
3) learned bias prediction for Invariant VIO (LBP-InVIO)~\cite{altawaitan2025learned}.
\begin{figure}[t]
    \centering
    \includegraphics[width=1.0\linewidth]{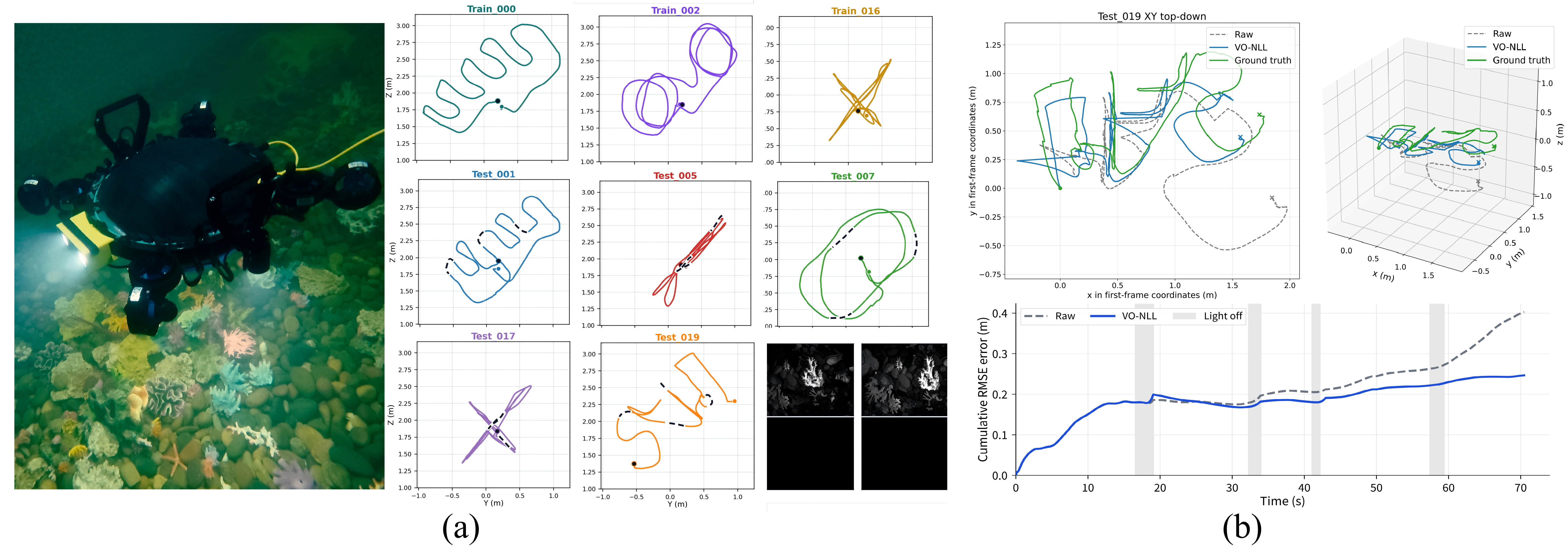}
    \small{\caption{
    (a) Overview of the \textsc{AquaLux} dataset. 
    (b) An OpenVINS testing example under intermittent visual outages. 
    The lower plot reports the cumulative position RMSE over time, showing that uncompensated IMU propagation accumulates larger errors after each light-off interval.
    }}
    \label{fig:aqua}
\end{figure}

\subsection{EuRoC Dataset Experiment}
The dataset EuRoC~\cite{burri2016euroc} provides 200 Hz IMU data and 200 Hz post-processed ground truth. 
The training and testing split follows the HL-CBD~\cite{liu2025debiasing}.
MSE initialization is performed for 1500 epochs, followed by 300 NLL epochs with alternating updates of the bias dynamics and covariance parameters. 
This paper adopts a scale parameterization
$Q(\psi)=\operatorname{diag}(\sigma_a^2 I_3,\sigma_g^2 I_3)$
to match the OpenVINS configuration, and the bias random-walk noises in OpenVINS are set to small values.
All proposed variants use a window size of 64 ATOs, corresponding to about 3.2~s and 640 IMU propagation steps for VO supervision at 20~Hz, and about 0.32~s for GT at 200~Hz. 

The results are reported in Table~1. 
LBP-InVIO learns a direct observation-to-bias mapping for invariant visual-inertial odometry, 
and is included to evaluate the benefit of dynamics-based bias modeling, with comparison limited to the reported EuRoC results for code unavailability.
In the proposed method, ATOs are instantiated by two sources: ground-truth~(GT) and stereo visual odometry~(VO) generated by ORB-SLAM3~\cite{9440682}.
Although VO estimates contain estimation noise and drift, our method outperforms HL-CBD, which is trained with GT supervision. 
This result suggests that the proposed methods can effectively exploit noisy trajectory observations.
As shown in Table~2, under the same GT supervision, an MSE-only variant is trained with the warm-up MSE objective for all 1800 epochs, without switching to the NLL stage. The MSE-to-NLL training reduces the average position error, indicating the benefit of the likelihood-based learning objective.
\begin{table}[H]
\centering
\small{\caption{EuRoC test-set results for raw integration and OpenVINS. Best and second-best results are marked in \textbf{bold} and \underline{underlined}, respectively.
The supervision source is shown in parentheses.}}
\scriptsize
\setlength{\tabcolsep}{2.8pt}
\renewcommand{\arraystretch}{1.12}
\resizebox{\textwidth}{!}{
\begin{tabular}{lccc cccc}
\toprule
\multirow{2}{*}{Seq.}
& \multicolumn{3}{c}{Raw Integration~(AOE$\downarrow$~(deg) / APE$\downarrow$~(m))}
& \multicolumn{4}{c}{OpenVINS~(AOE$\downarrow$~(deg) / APE$\downarrow$~(m))} \\
\cmidrule(lr){2-4}
\cmidrule(lr){5-8}
& (GT)~HL-CBD~\cite{liu2025debiasing}
& (GT)~Ours
& (VO)~Ours
& (GT)~HL-CBD~\cite{liu2025debiasing}
& (GT)~LBP-InVIO~\cite{altawaitan2025learned}
& (GT)~Ours
& (VO)~Ours \\
\midrule
MH02 (149.960 s)
& $1.410 / 161.816$
& $\best{1.027} / \best{85.002}$
& $\second{1.200} / \second{140.480}$
& $1.010 / 0.220$
& $-$
& $\second{0.734} / \second{0.083}$
& $\best{0.680} / \best{0.079}$ \\

MH04 (98.760 s)
& $1.677 / \best{92.715}$
& $\second{1.371} / \second{100.679}$
& $\best{1.331} / 106.841$
& $\second{0.650} / 0.310$
& $-$
& $\best{0.433} / \best{0.192}$
& $0.799 / \second{0.249}$ \\

V101 (115.450 s)
& $3.536 / 1486.906$
& $\second{3.177} / \best{1310.107}$
& $\best{3.075} / \second{1313.105}$
& $1.440 / 0.110$
& $-$
& $\second{0.568} / \second{0.044}$
& $\best{0.565} / \best{0.043}$ \\

V103 (104.655 s)
& $5.086 / 886.348$
& $\second{4.855} / \second{881.565}$
& $\best{4.682} / \best{855.191}$
& $\best{0.850} / \second{0.120}$
& $1.584 / 0.140$
& $\second{1.141} / \best{0.111}$
& $1.147 / \best{0.111}$ \\

V202 (144.700 s)
& $5.255 / 790.589$
& $\second{4.948} / \best{779.040}$
& $\best{4.894} / \second{804.737}$
& $2.600 / 0.160$
& $2.198 / 0.125$
& $\best{0.903} / \best{0.048}$
& $\second{0.919} / \second{0.050}$ \\
\midrule
Average
& $3.393 / 683.675$
& $\second{3.076} / \best{631.279}$
& $\best{3.036} / \second{644.071}$
& $1.310 / 0.180$
& $-$
& $\best{0.756} / \best{0.095}$
& $\second{0.822} / \second{0.106}$ \\
\bottomrule
\end{tabular}
}
\label{tab:euroc_raw_openvins_combined}
\end{table}

The peak GPU memory comparison between AD and the proposed double-adjoint implementation is shown in Table~2. 
AD memory increases rapidly with the integration window size, whereas double-adjoint memory is much less sensitive to the window size, enabling training over longer integration horizons.
This feature is also practically important for the next real-world underwater robot experiments, where onboard computers are typically resource-constrained.
\begin{table}[H]
\centering
\small{\caption{Left: OpenVINS ablation. Right: peak GPU memory for different window sizes.}}
\scriptsize
\setlength{\tabcolsep}{3.0pt}
\renewcommand{\arraystretch}{0.92}
\setlength{\aboverulesep}{0.2pt}
\setlength{\belowrulesep}{0.2pt}
\begin{tabular}{@{}lcc@{\hspace{12pt}} rcc@{\hspace{6pt}} rcc@{}}
\toprule
\multicolumn{3}{c@{\hspace{12pt}}}{Objective ablation}
& \multicolumn{6}{c}{Peak GPU memory$\downarrow$} \\
\cmidrule(lr){1-3}
\cmidrule(lr){4-9}
Metric & (GT)~MSE & (GT)~NLL
& $W$ & AD~(MB) & Double adjoint~(MB)
& $W$ & AD~(MB) & Double adjoint~(MB) \\
\midrule
AOE$\downarrow$~(deg) & 0.764 & \textbf{0.756}
& 16 & 525.6  & \textbf{351.3}
& 64  & 1069.3 & \textbf{279.8} \\
APE$\downarrow$~(m) & 0.102 & \textbf{0.095}
& 32 & 745.5  & \textbf{366.9}
& 128 & 2067.2 & \textbf{457.0} \\
\bottomrule
\end{tabular}
\label{tab:ablation_memory1}
\end{table}

\subsection{Application in Underwater VIO}
\textsc{AquaLux} is introduced as a benchmark for evaluating inertial odometry under underwater visual outages, as shown in~Figure~\ref{fig:aqua}~(a).
The controlled light-switching provides repeatable scenarios with temporary visual degradation caused by lighting failure, power instability, turbidity, occlusion, or low texture.
The dataset contains ten training sequences under reliable visual conditions, and ten test sequences with paired trajectory designs, but includes intermittent light-off intervals.
Eight sequences are used for training and five for testing in our experiment, covering both matched trajectory patterns and unseen trajectories~(Test\_005, Test\_007) for evaluating generalization.
The frequency of IMU is 1000 Hz, and the ground truth trajectory is provided by motion capture at 120 Hz. 
The training-stage VO has average position and rotation RMSEs of 0.043~m and 2.83~deg at 40 Hz. 
All methods are trained using a rollout window of 64 steps with a batch size of 1000.

\begin{table}[H]
\centering
\small{\caption{\textsc{AquaLux} test-set results for preintegration and OpenVINS. }}
\label{tab:aqualux_preintegration_openvins_combined}
\scriptsize
\setlength{\tabcolsep}{1.8pt}
\renewcommand{\arraystretch}{0.92}
\setlength{\aboverulesep}{0.15pt}
\setlength{\belowrulesep}{0.15pt}
\renewcommand{\arraystretch}{1.12}
\resizebox{\textwidth}{!}{
\begin{tabular}{lc ccc ccc}
\toprule
\multirow{2}{*}{Seq.}
& \multirow{2}{*}{Light-off ratio~(\%)}
& \multicolumn{3}{c}{Preintegration~(Rot. RMSE$\downarrow$~(deg) / Pos. RMSE$\downarrow$~(m))}
& \multicolumn{3}{c}{OpenVINS~(Rot. RMSE$\downarrow$~(deg) / Pos. RMSE$\downarrow$~(m))} \\
\cmidrule(lr){3-5}
\cmidrule(lr){6-8}
& 
& Raw
& (GT)~HL-CBD~\cite{liu2025debiasing}
& (VO)~Ours
& Raw
& HL-CBD~\cite{liu2025debiasing}
& (VO)~Ours \\
\midrule
Test\_001
& $10.427$
& $36.991 / 2362.635$
& $3.275 / 187.850$
& $\best{3.161} / \best{152.040}$
& $8.456 / 0.854$
& $3.976 / \best{0.145}$
& $\best{3.185} / 0.149$ \\

Test\_005
& $10.295$
& $39.327 / 2715.495$
& $3.789 / 204.714$
& $\best{2.988} / \best{168.111}$
& $\best{4.443} / 0.749$
& $5.285 / \best{0.251}$
& $5.711 / 0.264$ \\

Test\_007
& $9.276$
& $39.610 / 2788.245$
& $\best{6.182} / 472.936$
& $6.213 / \best{436.972}$
& $10.635 / 7.339$
& $\best{10.000} / \best{0.134}$
& $10.880 / 0.180$ \\

Test\_017
& $8.562$
& $46.918 / 4273.670$
& $4.849 / 499.592$
& $\best{3.357} / \best{444.660}$
& $4.479 / 0.177$
& $4.119 / 0.126$
& $\best{3.232} / \best{0.106}$ \\

Test\_019
& $10.366$
& $41.295 / 2952.144$
& $5.308 / 409.265$
& $\best{4.160} / \best{363.689}$
& $18.408 / 0.220$
& $49.239 / 1.238$
& $\best{8.115} / \best{0.158}$ \\
\midrule
Average
& $9.785$
& $40.828 / 3018.438$
& $4.681 / 354.871$
& $\best{3.976} / \best{313.094}$
& $9.284 / 1.868$
& $14.524 / 0.379$
& $\best{6.224} / \best{0.171}$ \\
\bottomrule
\end{tabular}
}
\end{table}
The experimental results are shown in Table~3. 
Directly integrating the IMU raw measurements performs poorly when visual sensing fails.
Although the proposed method uses noisier VO estimates as supervision, it still outperforms the GT-based HL-CBD on average. 
This is attributed to the longer time horizon covered by each rollout: 64 VO steps span about 1.6~s, while 64 IMU-rate steps span only 0.064~s. 
Thus, VO supervision exposes the model to longer inertial propagation intervals during training to capture longer-horizon dynamics.
We also find that short VO-supervision windows are less effective because the IMU preintegration error can be comparable to the VO error, making the supervisory signal less effective.
Moreover, the high-rate IMU in \textsc{AquaLux} amplifies drift from inaccurate bias compensation. 
Through inverse-covariance weighting, the likelihood downweights uncertain long-horizon residuals, preventing them from dominating optimization.
\section{Limitations}
\label{sec:limitations}
Several limitations remain. First, PLATO is trained offline, which avoids additional downstream runtime cost but limits immediate adaptation to platform changes, sensor aging, or environment-specific motion patterns. Visually reliable intervals in new sequences may support fine-tuning, while online adaptation is left for future work. Second, the covariance model uses a compact isotropic parameterization compatible with VIO systems such as OpenVINS. With noisy ATOs, the learned covariance should be viewed as an effective likelihood weight rather than a fully calibrated physical sensor covariance. More expressive covariance models may better capture direction-dependent or time-varying uncertainty. Finally, broader evaluations on more datasets, platforms, and environmental conditions are needed to more thoroughly assess generalization.

\section{Conclusion}
\label{sec:conclusion}
This paper presented PLATO, a likelihood-based framework for ATO-supervised neural inertial preintegration. PLATO refines the IMU measurement model by jointly learning NODE-based bias dynamics and IMU noise statistics under a unified marginal NLL objective.
The framework exploits the sparse information structure of invariant-error preintegration for efficient likelihood evaluation and gradient computation. For bias-dynamics training, the double-adjoint scheme couples a discrete invariant-error adjoint with a continuous-time bias adjoint, avoiding direct reverse-mode AD through the full nested bias-dynamics and IMU-preintegration rollout. Experiments on EuRoC show improved raw preintegration and downstream OpenVINS performance, with VO-supervised NLL comparable to GT-supervised training despite VO noise and drift. Validation on \textsc{AquaLux} demonstrates effectiveness for underwater VIO under intermittent visual degradation. Future work will investigate online adaptation and more expressive covariance models.


\clearpage


\clearpage

\bibliography{corlref}  

@article{barrau2016invariant,
  title={The invariant extended Kalman filter as a stable observer},
  author={Barrau, Axel and Bonnabel, Silv{\`e}re},
  journal={IEEE Transactions on Automatic Control},
  volume={62},
  number={4},
  pages={1797--1812},
  year={2016},
  publisher={IEEE}
}

@article{hartley2020contact,
  title={Contact-aided invariant extended Kalman filtering for robot state estimation},
  author={Hartley, Ross and Ghaffari, Maani and Eustice, Ryan M and Grizzle, Jessy W},
  journal={The International Journal of Robotics Research},
  volume={39},
  number={4},
  pages={402--430},
  year={2020},
  publisher={Sage Publications Sage UK: London, England}
}

@inproceedings{liu2025debiasing,
  title={Debiasing 6-DoF IMU via Hierarchical Learning of Continuous Bias Dynamics},
  author={Liu, Ben and Lin, Tzu-Yuan and Zhang, Wei and Ghaffari, Maani},
  booktitle={Robotics: Science and Systems (RSS)},
  year={2025}
}

@article{liu2020tlio,
  title={Tlio: Tight learned inertial odometry},
  author={Liu, Wenxin and Caruso, David and Ilg, Eddy and Dong, Jing and Mourikis, Anastasios I and Daniilidis, Kostas and Kumar, Vijay and Engel, Jakob},
  journal={IEEE Robotics and Automation Letters},
  volume={5},
  number={4},
  pages={5653--5660},
  year={2020},
  publisher={IEEE}
}

@inproceedings{herath2020ronin,
  title={Ronin: Robust neural inertial navigation in the wild: Benchmark, evaluations, \& new methods},
  author={Herath, Sachini and Yan, Hang and Furukawa, Yasutaka},
  booktitle={2020 IEEE international conference on robotics and automation (ICRA)},
  pages={3146--3152},
  year={2020},
  organization={IEEE}
}

@inproceedings{zhou2025learning,
  title={Learning imu bias with diffusion model},
  author={Zhou, Shenghao and Katragadda, Saimouli and Huang, Guoquan},
  booktitle={2025 IEEE International Conference on Robotics and Automation (ICRA)},
  pages={2162--2168},
  year={2025},
  organization={IEEE}
}

@inproceedings{sun2021idol,
  title={IDOL: Inertial deep orientation-estimation and localization},
  author={Sun, Scott and Melamed, Dennis and Kitani, Kris},
  booktitle={Proceedings of the AAAI Conference on Artificial Intelligence},
  volume={35},
  number={7},
  pages={6128--6137},
  year={2021}
}

@article{brossard2020denoising,
  title={Denoising IMU gyroscopes with deep learning for open-loop attitude estimation},
  author={Brossard, Martin and Bonnabel, Silvere and Barrau, Axel},
  journal={IEEE Robotics and Automation Letters},
  volume={5},
  number={3},
  pages={4796--4803},
  year={2020},
  publisher={IEEE}
}

@article{qiu2023airimu,
  title={Airimu: Learning uncertainty propagation for inertial odometry},
  author={Qiu, Yuheng and Wang, Chen and Xu, Can and Chen, Yutian and Zhou, Xunfei and Xia, Youjie and Scherer, Sebastian},
  journal={arXiv preprint arXiv:2310.04874},
  year={2023}
}

@article{altawaitan2025learned,
  title={Learned IMU Bias Prediction for Invariant Visual Inertial Odometry},
  author={Altawaitan, Abdullah and Stanley, Jason and Ghosal, Sambaran and Duong, Thai and Atanasov, Nikolay},
  journal={IEEE Robotics and Automation Letters},
  year={2025},
  publisher={IEEE}
}

@inproceedings{buchanan2022learning,
  title={Learning inertial odometry for dynamic legged robot state estimation},
  author={Buchanan, Russell and Camurri, Marco and Dellaert, Frank and Fallon, Maurice},
  booktitle={Conference on robot learning},
  pages={1575--1584},
  year={2022},
  organization={PMLR}
}

@article{jayanth2025neural,
  title={Neural Inertial Odometry from Lie Events},
  author={Jayanth, Royina Karegoudra and Xu, Yinshuang and Chatzipantazis, Evangelos and Daniilidis, Kostas and Gehrig, Daniel},
  journal={arXiv preprint arXiv:2505.09780},
  year={2025}
}

@article{chen2018neural,
  title={Neural ordinary differential equations},
  author={Chen, Ricky TQ and Rubanova, Yulia and Bettencourt, Jesse and Duvenaud, David K},
  journal={Advances in neural information processing systems},
  volume={31},
  year={2018}
}

@article{kidger2020neural,
  title={Neural controlled differential equations for irregular time series},
  author={Kidger, Patrick and Morrill, James and Foster, James and Lyons, Terry},
  journal={Advances in neural information processing systems},
  volume={33},
  pages={6696--6707},
  year={2020}
}

@inproceedings{geneva2020openvins,
  title={Openvins: A research platform for visual-inertial estimation},
  author={Geneva, Patrick and Eckenhoff, Kevin and Lee, Woosik and Yang, Yulin and Huang, Guoquan},
  booktitle={2020 IEEE International Conference on Robotics and Automation (ICRA)},
  pages={4666--4672},
  year={2020},
  organization={IEEE}
}

@ARTICLE{10757429,
  author={Zheng, Chunran and Xu, Wei and Zou, Zuhao and Hua, Tong and Yuan, Chongjian and He, Dongjiao and Zhou, Bingyang and Liu, Zheng and Lin, Jiarong and Zhu, Fangcheng and Ren, Yunfan and Wang, Rong and Meng, Fanle and Zhang, Fu},
  journal={IEEE Transactions on Robotics}, 
  title={FAST-LIVO2: Fast, Direct LiDAR–Inertial–Visual Odometry}, 
  year={2025},
  volume={41},
  number={},
  pages={326-346},
  doi={10.1109/TRO.2024.3502198}}

@article{buchanan2022deep,
  title={Deep imu bias inference for robust visual-inertial odometry with factor graphs},
  author={Buchanan, Russell and Agrawal, Varun and Camurri, Marco and Dellaert, Frank and Fallon, Maurice},
  journal={IEEE Robotics and Automation Letters},
  volume={8},
  number={1},
  pages={41--48},
  year={2022},
  publisher={IEEE}
}

@article{sola2018micro,
  title={A micro lie theory for state estimation in robotics},
  author={Sola, Joan and Deray, Jeremie and Atchuthan, Dinesh},
  journal={arXiv preprint arXiv:1812.01537},
  year={2018}
}

@ARTICLE{9440682,
  author={Campos, Carlos and Elvira, Richard and Rodríguez, Juan J. Gómez and M. Montiel, José M. and D. Tardós, Juan},
  journal={IEEE Transactions on Robotics}, 
  title={ORB-SLAM3: An Accurate Open-Source Library for Visual, Visual–Inertial, and Multimap SLAM}, 
  year={2021},
  volume={37},
  number={6},
  pages={1874-1890},
  doi={10.1109/TRO.2021.3075644}}

@misc{grupp2017evo,
  title={evo: Python package for the evaluation of odometry and SLAM.},
  author={Grupp, Michael},
  howpublished={\url{https://github.com/MichaelGrupp/evo}},
  year={2017}
}

@article{burri2016euroc,
  title={The EuRoC micro aerial vehicle datasets},
  author={Burri, Michael and Nikolic, Janosch and Gohl, Pascal and Schneider, Thomas and Rehder, Joern and Omari, Sammy and Achtelik, Markus W and Siegwart, Roland},
  journal={The International Journal of Robotics Research},
  volume={35},
  number={10},
  pages={1157--1163},
  year={2016},
  publisher={SAGE Publications Sage UK: London, England}
}

@inproceedings{barfoot2014batch,
  title={Batch Continuous-Time Trajectory Estimation as Exactly Sparse Gaussian Process Regression.},
  author={Barfoot, Tim D and Tong, Chi Hay and S{\"a}rkk{\"a}, Simo},
  booktitle={Robotics: Science and Systems},
  volume={10},
  pages={1--10},
  year={2014},
  organization={Citeseer}
}

@article{li2022closed,
  title={Closed-Form Error Propagation on $ SE\_ $\{$n$\}$(3) $ Group for Invariant EKF With Applications to VINS},
  author={Li, Xinghan and Jiang, Haodong and Chen, Xingyu and Kong, He and Wu, Junfeng},
  journal={IEEE Robotics and Automation Letters},
  volume={7},
  number={4},
  pages={10705--10712},
  year={2022},
  publisher={IEEE}
}
\appendix

\clearpage
\begin{center}
    {\Large \bfseries Appendix}
\end{center}

\section{Details of the \textsc{AquaLux} Dataset}
\begin{figure}[h]
    \centering
    \includegraphics[width=0.9\linewidth]{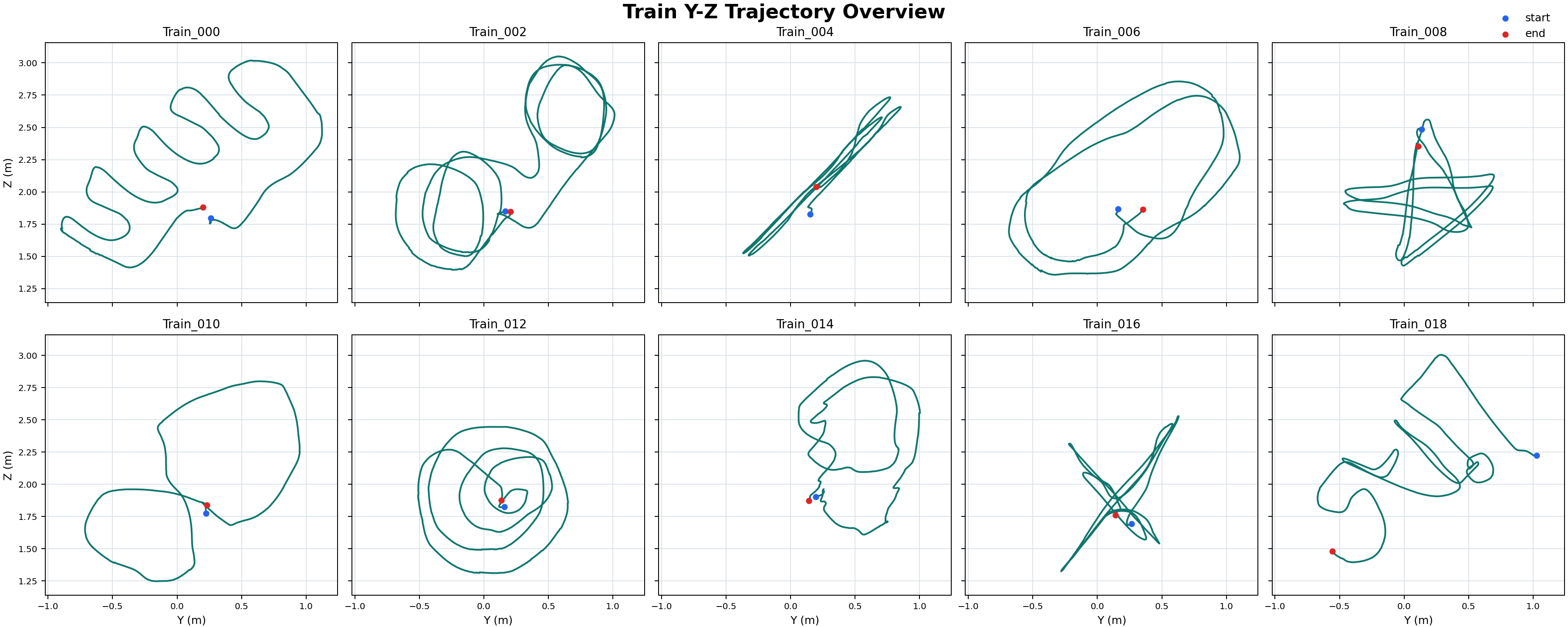}
    \caption{Training sequences provided in \textsc{AquaLux}.}
    \label{fig:all-train}
\end{figure}

\begin{figure}[h]
    \centering
    \includegraphics[width=0.9\linewidth]{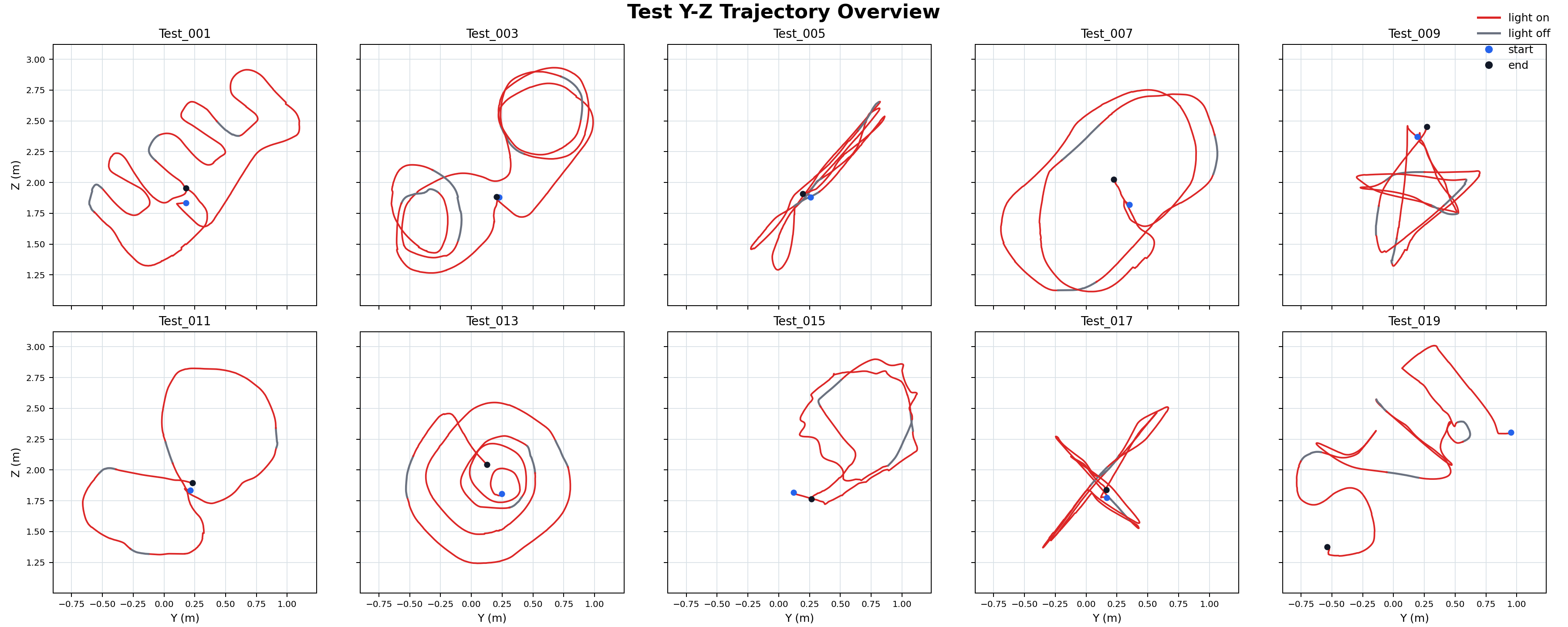}
    \caption{Testing sequences provided in \textsc{AquaLux}.}
    \label{fig:all-test}
\end{figure}

All training and testing sequences in the \textsc{AquaLux} dataset are shown in Figure \ref{fig:all-train} and~\ref{fig:all-test}. 

\begin{figure}[H]
    \centering
    \includegraphics[width=0.7\linewidth]{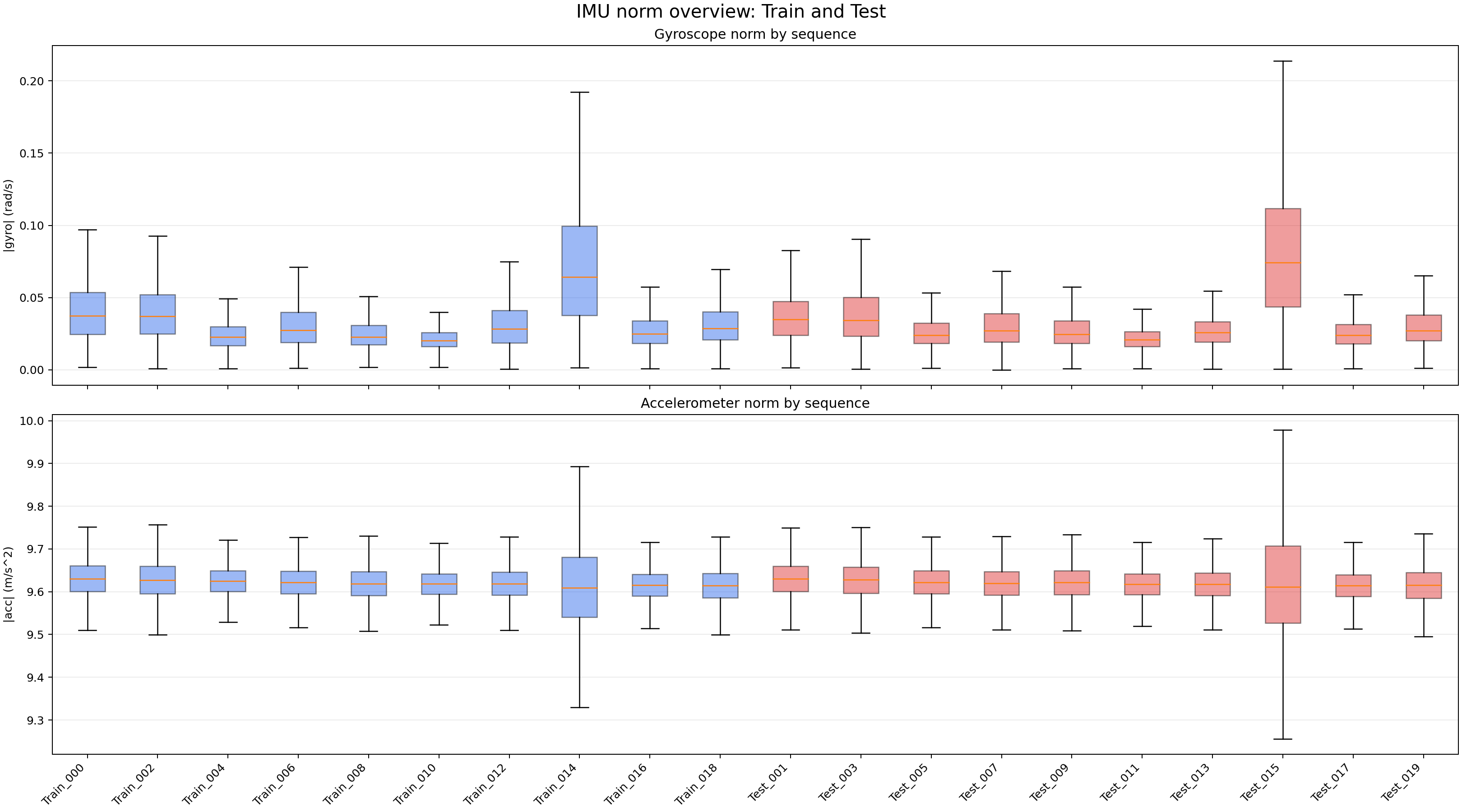}
    \caption{IMU data visualization.}
    \label{fig:imu}
\end{figure}

The visualization of IMU data norm is shown in Figure~\ref{fig:imu}.

\section{Sparse Negative Log-Likelihood}
\label{sec:sparse_nll}

This appendix derives the sparse information-form evaluation of the NLL.
The supervised indices are $\{s_1,\ldots,s_M\}$.
Define \(\bm\xi^s \triangleq \operatorname{vec}(\xi_{s_1},\ldots,\xi_{s_M})\) with the corresponding sampled time $t_{s_i}$ for index $s_i$.
The goal is to evaluate
\begin{equation}\label{eq:nll_app}
        \mathcal L_{\mathrm{NLL}}(\Theta)
    =
    \frac12\bm r^\top\bm S^{-1}\bm r
    +
    \frac12\log\det\bm S,
    \qquad
    \bm S=\bm H\bar{\bm P}^s\bm H^\top+\bm W,
\end{equation}
using sparse $\bar{\bm{\Lambda}}^s = (\bar{\bm P}^s)^{-1}$ without forming the dense covariance \(\bar{\bm P}^s\).

\paragraph{Sparse information matrix induced by preintegration.}
For \(r\ge q\), define the state-transition matrix between two IMU time indices as
$
    \Phi_{r\leftarrow q}
    \triangleq
    F_{r-1}F_{r-2}\cdots F_q,~
    \Phi_{q\leftarrow q}=I .
$
For two consecutive retained states \(X_{s_i}\) and \(X_{s_{i+1}}\), define
\(
    \Phi_i^s
    \triangleq
    \Phi_{s_{i+1}\leftarrow s_i}
\)
and the accumulated process covariance
\begin{equation}\label{eq:Qis}
    Q_i^s
    \triangleq
    \sum_{k=s_i}^{s_{i+1}-1}
    \Phi_{s_{i+1}\leftarrow k+1}
    G_k Q(\psi)G_k^\top
    \Phi_{s_{i+1}\leftarrow k+1}^\top .    
\end{equation}
Let \(P_1^s\succ0\) be the prior covariance of \(\xi_{s_1}\). 
The nonzero blocks of the block-tridiagonal precision matrix
\(\bar{\bm\Lambda}^s\) are given by
\[
\begin{aligned}
    \bar{\Lambda}_{11}^s
    &=
    (P_1^s)^{-1}
    +
    (\Phi_1^s)^\top(Q_1^s)^{-1}\Phi_1^s, \quad \bar{\Lambda}_{MM}^s=
    (Q_{M-1}^s)^{-1}, \\
    \bar{\Lambda}_{ii}^s
    &=
    (Q_{i-1}^s)^{-1}
    +
    (\Phi_i^s)^\top(Q_i^s)^{-1}\Phi_i^s, \quad i=2,\ldots,M-1, \\
    \bar{\Lambda}_{i,i+1}^s
    &=
    -(\Phi_i^s)^\top(Q_i^s)^{-1},
    \quad
    \bar{\Lambda}_{i+1,i}^s
    =
    -(Q_i^s)^{-1}\Phi_i^s,
    \quad i=1,\ldots,M-1 .
\end{aligned}
\]
According to Theorem~1 in~\cite{barfoot2014batch}, the inverse covariance of a Markov Gaussian trajectory prior is exactly sparse, and
$\bar{\bm \Lambda}^s=\bm B^{-\top} \bm{\Omega}^s \bm B^{-1}$, where
\[
    \bm{\Omega}^s
    =
    \left[
    \begin{smallmatrix}
        (P_1^s)^{-1} & 0 & \cdots & 0 \\
        0 & (Q_1^s)^{-1} & \cdots & 0 \\
        \vdots & \vdots & \ddots & \vdots \\
        0 & 0 & \cdots & (Q_{M-1}^s)^{-1}
    \end{smallmatrix}
    \right],
    \quad
    \bm B^{-1}
    =
    \left[
    \begin{smallmatrix}
        I & 0 & 0 & \cdots & 0 \\
        -\Phi_1^s & I & 0 & \cdots & 0 \\
        0 & -\Phi_2^s & I & \cdots & 0 \\
        \vdots & \vdots & \ddots & \ddots & \vdots \\
        0 & 0 & \cdots & -\Phi_{M-1}^s & I
    \end{smallmatrix}
    \right].
\]
Accordingly,  
$
    -\log\det\bar{\bm\Lambda}^s
    =
    \log\det P_1^s
    +
    \sum_{i=1}^{M-1}\log\det Q_i^s .
$

\paragraph{Sensitivity with respect to \(\psi\).}
During the optimization of \(\psi\), the bias-dynamics parameters \(\theta\) are fixed.
Consequently, the nominal states, the residuals, and the linearized propagation
matrices \(F_k\) and \(G_k\) are fixed.
The dependence of \(\mathcal L_{\mathrm{NLL}}\) on \(\psi\) is therefore only
through the IMU noise covariance \(Q(\psi)\), the induced interval covariances
\(Q_i^s(\psi)\), and the corresponding precision blocks in
\(\bar{\bm\Lambda}^s\).
For each supervisory interval \([s_i,s_{i+1}]\), the interval covariance~\eqref{eq:Qis} can be
computed by a local covariance recursion
\begin{equation}\label{eq:Qs_recursion}
        \Sigma_{i,k+1}
    =
    F_k\Sigma_{i,k}F_k^\top
    +
    G_kQ(\psi)G_k^\top,
    \qquad k=s_i,\ldots,~s_{i+1}\!-\!1,
\end{equation}
where \(F_k\triangleq\Phi_{k+1\leftarrow k}\), \(\Sigma_{i,s_i}=\bm 0\) and the recursion gives that \(Q_i^s=\Sigma_{i,s_{i+1}}\).
Then differentiating~\eqref{eq:Qs_recursion} term by term, follows that 
\begin{equation}\label{eq:dQs_recursion}
    \frac{\partial \Sigma_{i,k+1}}{\partial \psi_j}
    =
    F_k
    \frac{\partial \Sigma_{i,k}}{\partial \psi_j}
    F_k^\top
    +
    G_k
    \frac{\partial Q(\psi)}{\partial \psi_j}
    G_k^\top,
    \qquad k=s_i,\ldots,~s_{i+1}\!-\!1,
\end{equation}
where $    \frac{\partial \Sigma_{i,s_i}}{\partial \psi_j}=\bm 0,
    \;
    \frac{\partial Q_i^s}{\partial \psi_j}
    =
    \frac{\partial \Sigma_{i,s_{i+1}}}{\partial \psi_j}$.

\paragraph{Noise-free full-state supervision.}
Let $r_i\triangleq \log(X_{s_i} \bar{X}_{s_i}^{-1})$ and the corresponding Jacobian is $H_i=-\operatorname{dexp}^{-1}_{-r_i}$.
For the ground-truth case,
\(\bm W=\bm 0\) and the full-state residual Jacobian
\(\bm H=\operatorname{blkdiag}(H_1,\ldots,H_M)\) is square and locally
invertible. Thus
$
    \bm S=\bm H\bar{\bm P}^s\bm H^\top .
$
Define $\tilde{\bm r}
    \triangleq
    \bm H^{-1}\bm r$, and since
$
    (\bm H\bar{\bm P}^s\bm H^\top)^{-1}
    =
    \bm H^{-\top}\bar{\bm\Lambda}^s\bm H^{-1},
    \;
    \log\det(\bm H\bar{\bm P}^s\bm H^\top)
    =
    -\log\det\bar{\bm\Lambda}^s
    +
    2\log|\det\bm H|,
$
the NLL becomes
\[
    \mathcal L_{\mathrm{NLL}}^{\mathrm{gt}}(\Theta)
    =
    \frac12\tilde{\bm r}^\top\bar{\bm\Lambda}^s\tilde{\bm r}
    -
    \frac12\log\det\bar{\bm\Lambda}^s
    +
    \log|\det\bm H| .
\]
Since \(\bar{\bm\Lambda}^s\) is block tridiagonal,
\[
\boxed{\displaystyle
    \mathcal L_{\mathrm{NLL}}^{\mathrm{gt}}(\Theta)
    =
    \frac12
    \sum_{i=1}^{M}
    \tilde r_i^\top\bar{\Lambda}_{ii}^s\tilde r_i
    +
    \sum_{i=1}^{M-1}
    \tilde r_i^\top\bar{\Lambda}_{i,i+1}^s\tilde r_{i+1}
    -
    \frac12\log\det\bar{\bm\Lambda}^s
    +
    \log|\det\bm H| .
}
\]
Here
\(\log|\det\bm H|=\sum_{i=1}^{M}\log|\det H_i|\). Substituting the
factorized determinant above gives the equivalent sum form used in the main
text.

\paragraph{Noisy ATO supervision.}
For each supervised time \(t_{s_i}\), the residual is defined as \(r_i = \log(T_{s_i}\bar{T}_{s_i}^{-1})\), and the corresponding linearized Jacobian is \(H_i =-\operatorname{dexp}_{-r_i}^{-1} E\), with \(E=\bigl[\begin{smallmatrix} I_3 & 0 & 0\\ 0 & 0 & I_3 \end{smallmatrix}\bigr]\)
For noisy pose supervision, \(\bm W=\operatorname{blkdiag}(W_1,\ldots,W_M)\succ0\), the sparsity of NLL is well studied in~\cite{barfoot2014batch}.
The Woodbury identity gives
\[
    \bm S^{-1}
    =
    \bm W^{-1}
    -
    \bm W^{-1}\bm H
    \left(
        \bar{\bm\Lambda}^s
        +
        \bm H^\top\bm W^{-1}\bm H
    \right)^{-1}
    \bm H^\top\bm W^{-1}.
\]
Define
$
    \bm K
    \triangleq
    \bar{\bm\Lambda}^s+\bm H^\top\bm W^{-1}\bm H,~
    \bm\gamma
    \triangleq
    \bm H^\top\bm W^{-1}\bm r,~\bm K\bm z=\bm\gamma .$
Then the first term in~\eqref{eq:nll_app} is
\[
  \frac{1}{2}  \bm r^\top\bm S^{-1}\bm r
    =
    \frac{1}{2}  \bm r^\top\bm W^{-1}\bm r
    -
    \frac{1}{2}  \bm\gamma^\top\bm z .
\]
The second term in~\eqref{eq:nll_app} is given by the matrix determinant lemma yields
\[
   \frac{1}{2}   \log\det\bm S
    =
  \frac{1}{2}    \log\det\bm W
    +
 \frac{1}{2}     \log\det\bm K
    -
  \frac{1}{2}    \log\det\bar{\bm\Lambda}^s .
\]
As a consequence, 
\[
\boxed{\displaystyle
    \mathcal L_{\mathrm{NLL}}^{\mathrm{pose}}(\Theta)
    =
    \frac12\sum_{i=1}^{M} r_i^\top W_i^{-1}r_i
    -
    \frac12\bm\gamma^\top\bm z
    +
    \frac12\sum_{i=1}^{M}\log\det W_i
    +
    \frac12\log\det\bm K
    -
    \frac12\log\det\bar{\bm\Lambda}^s .
}
\]

For the evaluation of \(\bm z\) and \(\log\det\bm K\), since \(\bm H\) and
\(\bm W\) are block diagonal, define the local measurement 
information-weighted residual as
\[
    \gamma_i
    \triangleq
    H_i^\top W_i^{-1}r_i .
\]
The matrix \(\bm K\) is block tridiagonal, with blocks
\[
    K_{ii}
    =
    \bar{\Lambda}_{ii}^s+H_i^\top W_i^{-1}H_i,
    \qquad
    K_{i,i+1}
    =
    \bar{\Lambda}_{i,i+1}^s,
    \qquad
    K_{i+1,i}
    =
    K_{i,i+1}^\top .
\]
Thus \(\bm K\bm z=\bm\gamma\) can be solved by sparse block factorization.
For the log determinant, use the block LDL recursion
\(
    D_1=K_{11},\;
    D_i
    =
    K_{ii}
    -
    K_{i,i-1}D_{i-1}^{-1}K_{i-1,i},\;i=2,\ldots,M\) and hence, 
\(\log\det\bm K
    =
    \sum_{i=1}^{M}\log\det D_i\).
And the fully sparse sum form is as follows
\[
\boxed{
\begin{aligned}
    \mathcal L_{\mathrm{NLL}}^{\mathrm{pose}}(\Theta)
    ={}&
    \frac12\sum_{i=1}^{M} r_i^\top W_i^{-1}r_i
    -
    \frac12\sum_{i=1}^{M}\gamma_i^\top z_i
    +
    \frac12\sum_{i=1}^{M}\log\det W_i \\
    &+
    \frac12\sum_{i=1}^{M}\log\det D_i
    +
    \frac12\log\det P_1^s
    +
    \frac12\sum_{i=1}^{M-1}\log\det Q_i^s .
\end{aligned}
}
\]
All terms are obtained from the block residuals \(r_i\), the local linearized
measurement models \((H_i,W_i)\), and the preintegration quantities
\((P_1^s,\Phi_i^s,Q_i^s)\).

\section{Invariant Error Adjoint}
\begin{figure}[H]
    \centering
    \begin{minipage}[t]{0.48\linewidth}
        \centering
        \includegraphics[width=\linewidth]{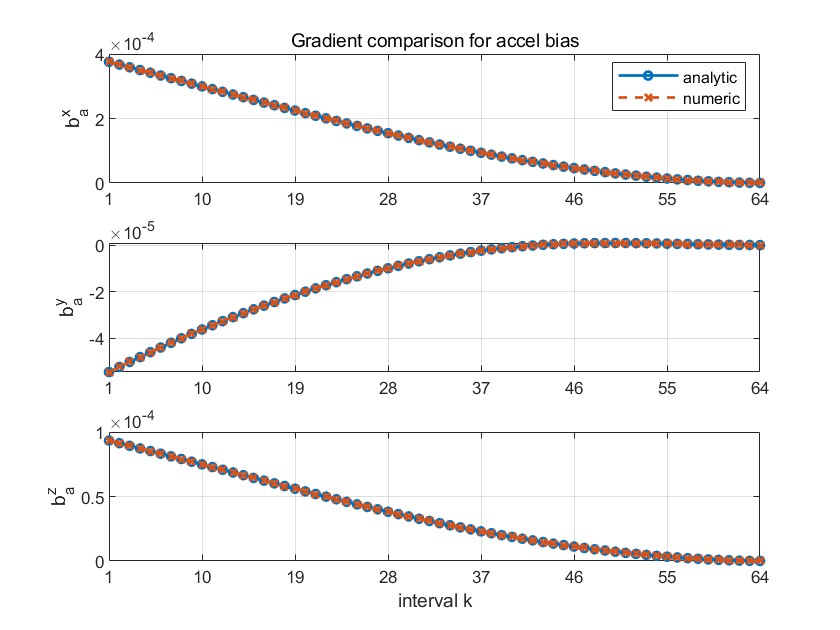}\\[-1mm]
        {\footnotesize (a) Accelerometer bias}
    \end{minipage}
    \hfill
    \begin{minipage}[t]{0.48\linewidth}
        \centering
        \includegraphics[width=\linewidth]{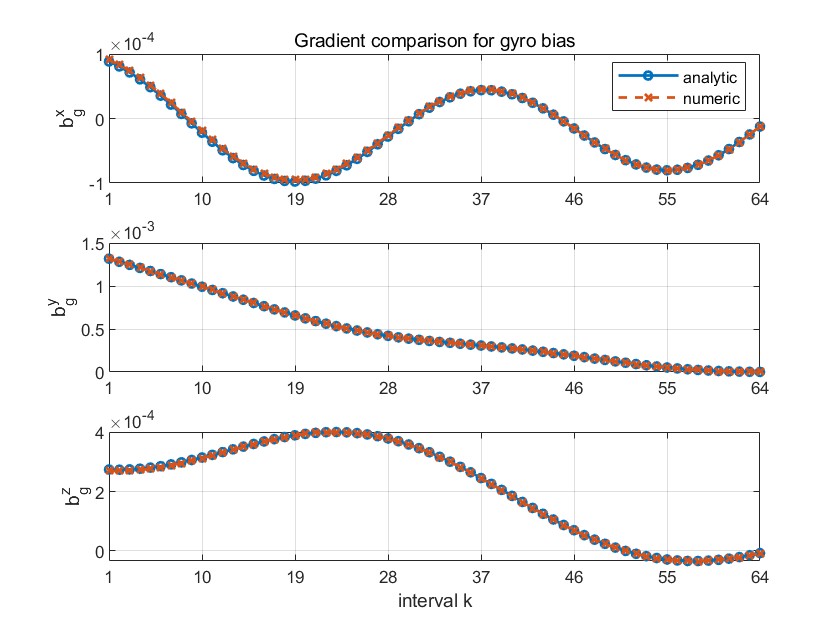}\\[-1mm]
        {\footnotesize (b) Gyroscope bias}
    \end{minipage}
    \caption{Numerical verification of the invariant-adjoint bias gradients.}
    \label{fig:invariant_adjoint_bias_verification}
\end{figure}

This section validates the outer invariant-error adjoint through numerical
simulation. Time-varying biases $b_t^a$ and $b_t^g$ are generated by
trigonometric functions, injected into the IMU measurement model, and used for
preintegration-based loss evaluation. The analytical gradient of
$\mathcal L$ with respect to $b(t)$ is compared with finite-difference gradients in Fig.~\ref{fig:invariant_adjoint_bias_verification}.

\end{document}